\documentclass[conference]{IEEEtran}
\usepackage{cite}
\usepackage{comment}
\usepackage{amsmath,amssymb,amsfonts}
\usepackage{float}
\usepackage{graphicx}
\usepackage[countmax]{subfloat}
\usepackage[caption=false,font=footnotesize]{subfig}
\usepackage{tabularx}
\usepackage{forest}
\usepackage{listings}
\usepackage{url}
\usepackage{hyperref}

\usepackage{pgfplots}
\usepackage{xcolor}
\usepackage{multirow}

\usepackage{threeparttable}

\pgfplotsset{compat=1.18}

\definecolor{precisionblue}{RGB}{52,152,219}
\definecolor{recallgreen}{RGB}{46,204,113}
\definecolor{accuracyyellow}{RGB}{241,196,15}
\definecolor{f1purple}{RGB}{155,89,182}

\lstdefinelanguage{json}{
    basicstyle=\ttfamily\footnotesize,
    numberstyle=\scriptsize,
    showstringspaces=false,
    breaklines=true,
    breakatwhitespace=false,
    frame=single,
    tabsize=2
}

\lstdefinestyle{prompt}{
    basicstyle=\ttfamily\small,
    breaklines=true,
    frame=single,
    showstringspaces=false,
    columns=flexible,
    keepspaces=true,
    breakindent=0pt,      
    resetmargins=true,    
    xleftmargin=0pt,      
    xrightmargin=0pt      
}

\begin{document}

\title{Schema-Guided Hierarchical Information Extraction and Semantic Evaluation Using Generative AI}

\author{
\IEEEauthorblockN{Modhurita Mitra\textsuperscript{a,1},
Jan-Willem Versteeg\textsuperscript{b,2},
Maarten D. Schermer\textsuperscript{a,3},\\
Shiva Nadi Najafabadi\textsuperscript{a,4},
Marie L. De Bruin\textsuperscript{b,5},
Lourens T. Bloem\textsuperscript{b,6}}
\IEEEauthorblockA{\textsuperscript{a}\textit{Research Engineering Team,
Information and Technology Services,}\\
\textit{Utrecht University, Utrecht, The Netherlands}}
\IEEEauthorblockA{\textsuperscript{b}\textit{Division of Pharmacoepidemiology
and Clinical Pharmacology,}\\
\textit{Utrecht Institute for Pharmaceutical Sciences,
Utrecht University, Utrecht, The Netherlands}}
\IEEEauthorblockA{\textsuperscript{1}m.mitra@uu.nl,
\textsuperscript{2}j.versteeg@uu.nl,
\textsuperscript{3}m.d.schermer@uu.nl,\\
\textsuperscript{4}s.nadinajafabadi@uu.nl,
\textsuperscript{5}m.l.debruin@uu.nl,
\textsuperscript{6}l.t.bloem@uu.nl}
}

\maketitle

\begin{abstract}
We present a schema-based framework for extracting complex, structured information from unstructured text documents using generative AI, followed by automated semantic evaluation of the extracted information against a gold standard. The schema, serving as an information model encoding domain knowledge, provides a unified, systematic, and consistent framework for extraction of hierarchical, nested information, with attributes of variable cardinality, and subsequent evaluation of the results. Information extraction from a document is performed in a single call to the model, in zero-shot mode.  

In the evaluation step, we introduce a path-based semantic matching algorithm to align the nested, variable-cardinality attributes in the extracted results with those in the gold standard. We use generative AI for semantic comparison of the extracted and gold standard values of an attribute, and introduce a rubric to classify the result of the comparison, according to domain-specific considerations, as an exact, semantic, useful, or non-match.  

We were able to extract 12 out of 14 attributes with an F1 score of $>$90\% from documents published by the health technology assessment organisation NICE, using the generative AI model Claude Opus 3. The time needed to extract the attributes from a document was $\sim$30 times lower than the time taken by a human domain expert. We further demonstrate generalisability of this framework across different generative AI models and transferability across different HTA organisations and languages. 
\end{abstract}


\maketitle

\section{Introduction}

The schema-based information extraction method that is the subject of this paper emerged from our efforts to extract a number of attributes of interest from documents from health technology assessment (HTA) organisations. The content of these documents is highly specialised and deep domain knowledge is required to interpret them, with the content and the structure of the documents differing from document to document, organisation to organisation, and over time. The structure of the desired attributes is complex and involves hierarchical relationships between attributes of variable cardinality (an attribute of variable cardinality is a list whose length is not known in advance). Extracting this kind of intricate, nuanced data from unstructured text is a difficult information extraction problem. 

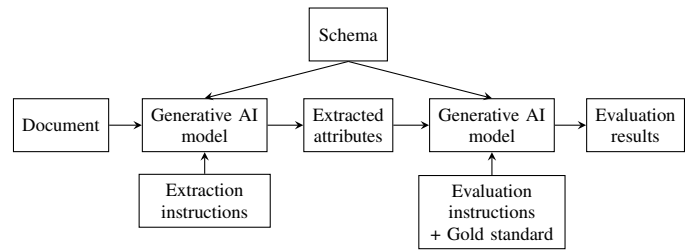
\begin{figure}[htbp]
\centering
\vspace{-5pt}
\begin{tikzpicture}[
    scale=1,
    transform shape,
    node distance=2.2cm,
    process/.style={rectangle, minimum width=1cm, minimum height=0.7cm, text centered, draw=black, align=center, font=\scriptsize},
    arrow/.style={->,>=stealth}
]
\node[process] (doc) {Document};
\node[process, right of=doc, xshift=-0.3cm] (genai1) {Generative AI\\model};
\node[process, right of=genai1, xshift=-0.3cm] (extracted) {Extracted\\attributes};
\node[process, right of=extracted, xshift=-0.3cm] (genai2) {Generative AI\\model};
\node[process, right of=genai2, xshift=-0.3cm] (results) {Evaluation\\results};

\node[process, above of=extracted, yshift=-1cm, text width=0.8cm] (schema) {Schema};

\node[process, below of=genai1, yshift=1.2cm, text width=1.5cm] (inst1) {Extraction instructions};
\node[process, below of=genai2, yshift=1.05cm, text width=1.7cm] (inst2) {Evaluation instructions \\+ Gold standard};

\draw[arrow] (doc) -- (genai1);
\draw[arrow] (genai1) -- (extracted);
\draw[arrow] (extracted) -- (genai2);
\draw[arrow] (genai2) -- (results);

\draw[arrow] (schema.south) -- (genai1.north);
\draw[arrow] (schema.south) -- (genai2.north);

\draw[arrow] (inst1) -- (genai1);
\draw[arrow] (inst2) -- (genai2);
\end{tikzpicture}
\caption{A high-level overview of the end-to-end information extraction and evaluation pipeline, highlighting the central role of the schema. A human-generated gold standard is needed for the evaluation step.}
\label{fig:extraction-pipeline}
\vspace{-10pt}
\end{figure}

In this proof-of-concept study carried out as a collaboration between HTA domain researchers and Research Engineers, we extracted these attributes using generative AI. We developed a unified schema-based framework to guide a generative AI model to perform information extraction as well as subsequent evaluation of the results. The schema serves as a single, shared information model \cite{sutherland1986model, ISO13972:2022, Wang2018} that encapsulates the domain knowledge needed for the information extraction, output formatting, and evaluation steps -- thus orchestrating the entire end-to-end information extraction and evaluation pipeline. Figure \ref{fig:extraction-pipeline} shows a schematic representation of the pipeline.

\section{Health Technology Assessment (HTA)}
\label{sec:hta}
Health Technology Assessment is the process of systematically reviewing a health technology (e.g., a drug, a medical device, or a medical therapy) based on clinical evidence summarised in reports that include conclusions regarding various attributes (e.g. clinical effectiveness and cost-effectiveness of drugs) that affect the adoption of that health technology in patient care \cite{orourke2020}. 

In the European Union, this process is largely carried out independently by each country, resulting in heterogeneous data in different languages and formats. To facilitate comparative HTA research, which is the comparison and analysis of HTA processes, practices, and recommendations across different countries and HTA organisations \cite{Vreman2020}, one needs to extract, for a given health technology, the same attributes from these different documents in a standardised format. Due to the complexity of HTA documents and the deep domain expertise required to interpret them, the current standard practice for extracting attributes from HTA documents is to employ domain experts to perform this information extraction task manually. 

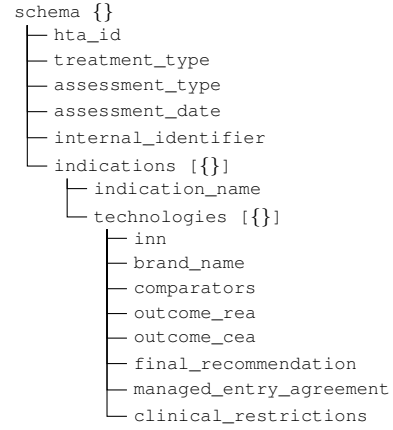
\begin{figure}[htbp]
\centering
\scriptsize
\begin{forest}
for tree={
  font=\ttfamily\scriptsize,
  grow'=0,
  child anchor=west,
  parent anchor=south west,
  anchor=west,
  calign=first,
  s sep=2pt,
  inner ysep=1pt,
  edge path={
    \noexpand\path [draw, \forestoption{edge}]
    (!u.south west) +(7.5pt,0) |- (.child anchor)\forestoption{edge label};
  },
  before typesetting nodes={
    if n=1
      {insert before={[,phantom]}}
      {}
  },
  fit=band,
  before computing xy={l=15pt},
}
[schema \{\}
  [hta\_id]
  [treatment\_type]
  [assessment\_type]
  [assessment\_date]
  [internal\_identifier]
  [indications \text{[\{\}]}
    [indication\_name]
    [technologies \text{[\{\}]}
      [inn]
      [brand\_name]
      [comparators]
      [outcome\_rea]
      [outcome\_cea]
      [final\_recommendation]
      [managed\_entry\_agreement]
      [clinical\_restrictions]
    ]
  ]
]
\end{forest}
\caption{Schema structure for attributes to be extracted from HTA documents, illustrating the hierarchical nature of the desired attributes. Here $\{\}$ denotes a dictionary and $\left[\right]$ denotes a list.}
\label{fig:schema}
\vspace{-10pt}
\end{figure}

The scientific goal of this project is to create an Open Science database of a number of attributes of interest extracted from these documents in order to facilitate their downstream use by different stakeholders. We identified 14 research-relevant attributes, listed in Table \ref{tab:hta_data_points}, that we wanted to extract from HTA documents for the purpose of comparative HTA research. The number of instances of each attribute can vary from document to document -- for example, multiple drugs might be assessed in a document, or one drug might be assessed for treating multiple indications (medical conditions). These attributes have a nested structure -- for example, each drug in the document has its own corresponding brand name, relative effectiveness assessment, and final recommendation. The nested structure of the attributes is illustrated in Figure \ref{fig:schema}. Our task was to extract these attributes from a set of HTA documents.

\begin{table}[htbp]
\centering
\vspace{-5pt}
\scriptsize
\caption{Attributes to be extracted from HTA documents}
\label{tab:hta_data_points}
    \centering
\begin{tabularx}{\columnwidth}{|
  >{\raggedright\arraybackslash\hsize=.6\hsize}X | 
  >{\raggedright\arraybackslash\hsize=1.4\hsize}X | 
}
\hline
\textbf{Attribute} & \textbf{Description} \\
\hline
HTA ID & Name of HTA organisation performing the assessment \\
Treatment type & Is the technology being assessed a medicine (drug), device, or therapy? \\
Assessment type & Is this the first assessment, a reassessment, or an indication broadening? \\
Assessment date & When was the assessment finalised/published? \\
Internal identifier & Code or label identifying the document \\
Indication & Medical condition for which the drug is assessed \\
INN & International nonproprietary name of assessed drug \\
Brand name & Brand name of assessed drug \\
Comparators & Drug(s) with which the performance of the assessed drug is compared \\
Relative effectiveness assessment outcome & Outcome of the relative effectiveness assessment for this drug-indication combination \\
Cost-effectiveness assessment outcome & Outcome of the cost-effectiveness assessment for this drug-indication combination \\
Final recommendation & What is the final reimbursement recommendation for this drug-indication combination? \\
Managed entry agreement & Was any managed entry agreement proposed? \\
Clinical restrictions & Clinical restrictions stated in the recommendation \\
\hline
\end{tabularx}
\vspace{-10pt}
\end{table}

\section{Related work}
\label{sec:related_work}

Information extraction is the extraction of structured data in machine-readable format from unstructured text data in digital format (e.g., \cite{Cowie1996, Sarawagi2008, Piskorski2013}). The main approaches to perform information extraction tasks have traditionally fallen into one of two categories: rule-based methods, and machine learning-based methods \cite{Wang2018, Xu2024}. 

With the advent of large language models (LLMs), these models are increasingly being used to perform information extraction tasks. To improve performance for certain tasks in specialised fields, fine-tuning BERT \cite{devlin-etal-2019-bert} or BART \cite{lewis-etal-2020-bart} models and, more recently, open-weights generative AI models, with domain-specific data has become popular practice \cite{Chen2025}. These fine-tuning methods require labelled training data, the technical know-how for fine-tuning LLMs, and the infrastructure to fine-tune and deploy the models. Thus this method is not easily accessible to domain experts with limited technical knowledge and infrastructural resources. 

On the other hand, there has been increasing interest in the development of novel prompting techniques to harness the language capabilities and knowledge inherent in generative AI models. These techniques make it possible to use general-purpose generative AI models for a wide range of natural language processing tasks, including information extraction. Some such prompt engineering techniques are chain-of-thought prompting \cite{Wei2022}, LLM-as-a-judge \cite{Zheng2023}, and zero-, few-, and many-shot prompting \cite{Kojima2022, Brown2020, Agarwal2024}. 

Structured information extraction via constrained decoding has become common practice in the past couple of years \cite{willard2023efficientguidedgenerationlarge, pokrass2024structuredoutputs, anthropic_structured_outputs_blog, geng2025jsonschemabenchrigorousbenchmarkstructured}. In constrained decoding, a response schema is passed to a generative AI API along with but separate from the prompt, and the AI produces a response consistent with the schema. 

LLMs fine-tuned on domain-specific materials science data have been used in a study to extract hierarchical attributes of variable cardinality  \cite{Dagdelen2024}. The nesting was only one level deep in this study, and only an exact string match was performed during automated evaluation.

In a study on populating knowledge bases, a schema for representing nested, hierarchical information was populated by recursively traversing it, and extracting the relevant information at each level by issuing a new prompt to a generative AI model \cite{Caufield2024}. An ontology was used for grounding and normalisation of the extracted information. However, this kind of grounding is not an evaluation of the inherent correctness of the extracted information -- it is only validation of the existence of the extracted information in the ontology. 

In a study on information extraction from histopathology reports, a JSON schema was provided in the prompt to guide the extraction \cite{BALASUBRAMANIAN2025100521}. However, this JSON schema was flat -- hierarchical, nested data with attributes of variable cardinality were not extracted in this study. 

\section{Data}
\label{sec:data}
The data consists of a set of HTA documents downloaded from the website of the National Institute for Health and Care Excellence (NICE) \cite{NICE_TechAppraisal} in the United Kingdom. 

We chose 8 representative documents as our development set (also known as validation set in machine learning). In addition, we manually created a gold standard for this development set.

The development set consisted of the documents we used to craft and refine our prompts. The representative development set covered a diverse range of values of the attributes -- some documents were initial assessments while others were reassessments, some values of the relative effectiveness, cost-effectiveness, and final recommendation were positive while others were negative, etc. In some documents, multiple drugs were assessed for the treatment of one indication, while in others, one drug was assessed for the treatment of multiple indications. 

The test set consisted of 50 randomly chosen HTA documents, for which we again manually created a gold standard. We evaluated the performance of our method on this test set, against the gold standard. The gold standard for the test set contained a total of 57 indications, 60 technologies, and 837 instances of leaf-level attributes. A leaf-level attribute is one which does not have any further sub-attributes in the schema in Figure \ref{fig:schema}.

\section{Method}
\label{sec:method}

Using our schema-based framework, we extract complex data -- hierarchical, with attributes of variable cardinality -- in a single model call, in zero-shot mode, from entire documents (20-40 pages long), and subsequently perform automated semantic evaluation of the extracted attributes. By single model call we mean that we make only one API call per document to the generative AI model in the extraction step; we send a single prompt to the model and it returns a single output containing the entire information to be extracted from the document, all at once. By zero-shot we mean that we do not provide examples in the prompt for in-context learning \cite{Kojima2022, Brown2020}. We simply provide a schema which serves as an information model, as well as accompanying instructions, to a state-of-the-art generative AI model, Claude Opus 3 from Anthropic \cite{claude3modelcard, anthropic_claude3}, accessed via its public API \cite{Anthropic_API}.

\subsection{Extraction}

\subsubsection{Methodology development}

We initially used a Q\&A method with the generative AI model Claude Opus 3 from Anthropic to extract the attributes of interest, instructing the model to return the extracted values in JSON object format \cite{mitra2024ieee}. This simple, flat JSON object structure turned out to not be applicable in a general manner to all documents. Some documents assessed one drug for treating multiple indications, while others assessed multiple drugs for treating one indication. It became evident that we needed a nested JSON structure to represent this kind of hierarchical data structure with attributes of variable cardinality.

We observed that a model could interpret a well-designed JSON schema functioning as an information model, and faithfully extract information that was consistent with this schema. We provided the schema in the prompt -- providing a response schema separately to the API for structured extraction was not available for Claude models till 2025 \cite{anthropic_structured_outputs_blog}, while we had started on schema-based extraction for this work in early 2024. A schema in the prompt provides a soft guideline for the output structure, whereas a response schema enforces the structure as a hard constraint. Nevertheless, every JSON output we obtained was schema-compliant, though we provided the JSON schema only in the prompt.   

\subsubsection{Model choice and settings}
We chose to use the generative AI model Claude Opus 3 from Anthropic because it produced the most accurate and consistent results \cite{data_consistency_accuracy, software_accuracy_reliability, reproducibility2019, mitra2024ieee} among the models we tried at the time we performed the extraction part of this study (2024-2025). The other models we tried were the GPT 3.5 \cite{gpt3modelcard} and GPT 4 \cite{gpt4systemcard} series of models from OpenAI. 

The temperature parameter in a generative AI model controls the randomness and variability of the outputs \cite{Anthropic_Glossary}. To make the results as deterministic and reproducible as possible, we set the temperature parameter to zero. 

The code, prompts, schema, and results can be found in the GitHub repository\footnote{\url{https://github.com/UtrechtUniversity/hta-genai}} for this project.

\subsubsection{Prompt}
The prompt consisted of three parts. The first part was the data -- the text of the HTA document, which we obtained by using the PyPDF package \cite{pypdf-docs} to convert the PDF documents to text format. The second part consisted of the instructions, which included setting the AI's role, description of the task to be performed, and requirements regarding the output format. The third part was the schema, which specified the attributes to be extracted and the format for the output. 

\subsection{Schema as information model}
\label{section_schema}

The schema,\footnote{The schema can be found here: \url{https://github.com/UtrechtUniversity/hta-genai/blob/main/config/schema.json}} expressed in JSON Schema format, plays a central role in this work. It serves as an information model \cite{sutherland1986model, ISO13972:2022, Wang2018} encoding domain knowledge, and comes into play multiple times and in multiple ways during the extraction and evaluation processes:

\begin{enumerate}
\item \textit{Encoding domain information:} The leaf-level keys in the schema specify the names of the attributes to be extracted. The \texttt{description} field provides a description of the attribute to be extracted, and any other relevant domain knowledge about the attribute.

\hspace{5pt} Where applicable, the \texttt{type}, \texttt{pattern}, and \texttt{enum} fields provide constraints on the value corresponding to a given key in the output. The \texttt{type} field constrains the type of the output -- for example, to ``string" or ``null". The \texttt{pattern} field forces the value to follow a certain format, for example YYYY-MM-DD for dates. The \texttt{enum} field restricts the value of the output to one of a limited number of pre-specified values -- for example to ``positive", ``negative", or \texttt{null}. The \texttt{enum} field thus provides a value set \cite{ecqi_value_set_2025, Wang2018} accompanying the information model defined by the schema.  

\hspace{5pt} The \texttt{type} field is also used to specify the nested, hierarchical, variable cardinality structure of the data -- if the \texttt{type} of an attribute is set to ``array", that means that the attribute is a list of unknown length; if \texttt{type} is ``object", that indicates that the attribute is a dictionary with further sub-attributes. 

\item \textit{Specifying output format:} The schema specifies the desired
output format -- the format at the level of the individual
key-value pairs for the attributes, as well as the format of the entire JSON
object with the nested structure and attributes of variable cardinality.

\item \textit{Providing context during evaluation:} During evaluation, we check if the extracted values match the gold standard values. If the values are not an exact string match, we use generative AI to check if they match semantically. For this semantic comparison, we provide the attribute name and the associated \texttt{description} field from the schema to provide context about the values being compared. 
\end{enumerate}

Figure \ref{fig:assessment-type-schema} shows an excerpt from the full schema. It shows the entry for one of the attributes to be extracted -- \texttt{assessment\_type}.

\begin{figure}[htbp]
\centering
\scriptsize
\begin{lstlisting}[language=json]
"assessment_type": {
  "type": ["string", "null"],
  "enum": ["initial assessment", "reassessment", "indication broadening", "non-submission", "unknown", null],
  "description": "Type of assessment being performed"}
\end{lstlisting}
\caption{Entry for the field \texttt{assessment\_type} in the schema
}

\label{fig:assessment-type-schema}
\end{figure}

For creating the schema, the HTA domain experts specified the attributes to extract, and their hierarchical, nested structure -- they knew which attributes had variable cardinality and thus needed to have a list structure, and which attributes had further sub-attributes and thus needed to have a dictionary structure. They communicated this information to the Research Engineers who then designed the schema in JSON Schema format. The domain experts also provided the domain knowledge encapsulated in the schema, such as the information in the \texttt{description}, \texttt{pattern}, \texttt{type}, and \texttt{enum} fields.

\subsection{Evaluation}
\label{section:evaluation}

The results are evaluated against a human-generated gold standard. Both the extracted results and the gold standard are generated in the format prescribed by the schema.

The fact that the schema has a nested structure complicates the process of finding the corresponding elements in the extracted JSON object and the gold standard JSON object. Some non-leaf-level attributes such as \texttt{indications} and \texttt{technologies} are lists containing multiple elements. The number of elements in such a list in the extracted JSON object might be different from the number of elements in the corresponding list in the gold standard JSON object. Even if the same number of elements are present in such a list in both the extracted and gold standard JSON objects, the elements might be in a different order in each of these lists. Therefore, direct comparison of the extracted JSON object with the gold standard is a non-trivial task. 

Manually aligning the extracted output with the gold standard and then comparing the corresponding attributes was extremely tedious, especially during the prompt engineering phase when we had to check the complicated nested JSON output manually each time we refined our prompt. This motivated us to develop an automated evaluation method.

\subsubsection{Methodology}

\begin{figure}[htbp]
\centering
\begin{tikzpicture}[
    scale=0.6,
    transform shape,
    node distance=2cm,
    font=\large,
    startstop/.style={rectangle, rounded corners, minimum width=7cm, minimum height=1cm, text centered, draw=black},
    inout/.style={trapezium, trapezium left angle=70, trapezium right angle=110, inner sep=5pt, align=center, draw=black},
    process/.style={rectangle, minimum width=5cm, minimum height=0.7cm, text centered, draw=black, text width=5cm, align=center},
    decision/.style={diamond, minimum width=0cm, minimum height=0cm, text centered, draw=black, aspect=2, text width=2cm, inner sep=0pt, align=center},
    arrow/.style={->,>=stealth}
]

\node (step1) [inout] {Gold standard, extracted results};

\node (step2) [process, below of=step1, yshift=0.7cm] {Compare top-level attributes};

\node (step2a) [process, below of=step2, yshift=0.3cm] {Form best-matching 1:1 (gold standard, extracted results) indication pairs};

\node (step3) [process, below of=step2a, yshift=0.2cm] {Select \\ one indication pair};

\node (step4) [decision, below of=step3, yshift=-0.1cm] {Both indications non-null?};

\node (step4a) [process, below of=step4, yshift=-0.4cm] {Form best-matching 1:1 (gold standard, extracted results) technology pairs};

\node (step5) [process, below of=step4a, yshift=0.2cm] {Select \\ one technology pair};

\node (step5a) [decision, below of=step5, yshift=-0.1cm] {Both technologies non-null?};

\node (step6) [decision, below of=step5a, yshift=-0.9cm] {INNs/ \\ brand names \\match?};

\node (step7) [process, below of=step6, yshift=-0.4cm] {Compare technology-level attributes};

\node (step8) [decision, below of=step7, yshift=-0.1cm] {More technology pairs?};

\node (step9) [decision, below of=step8, yshift=-0.7cm] {More indication pairs?};

\node (step10) [inout, below of=step9, yshift=0cm] {Evaluation results};

\draw [arrow] (step1) -- (step2);
\draw [arrow] (step2) -- (step2a);
\draw [arrow] (step2a) -- (step3);
\draw [arrow] (step3) -- (step4);
\draw [arrow] (step4) -- node[anchor=west] {Yes} (step4a);
\draw [arrow] (step4a) -- (step5);
\draw [arrow] (step4.east) -- node[anchor=south] {No} ++ (3.0,0) |- (step9.east);
\draw [arrow] (step5) -- (step5a);
\draw [arrow] (step5a) -- node[anchor=west] {Yes} (step6);
\draw [arrow] (step5a.east) -- node[anchor=south] {No} ++(2.0,0) -- ++(0,-1) |- (step8.east);
\draw [arrow] (step6) -- node[anchor=west] {Yes} (step7);
\draw [arrow] (step6.east) -- node[anchor=south] {No} ++(1.0,0) -- ++(0,-1) |- (step8.north east);
\draw [arrow] (step7) -- (step8);
\draw [arrow] (step8.west) -- node[anchor=south] {Yes} ++(-1,0) |- (step5.west);
\draw [arrow] (step8) -- node[anchor=west] {No} (step9);
\draw [arrow] (step9.west) -- node[above] {Yes} ++(-2,0) |- (step3.west);
\draw [arrow] (step9) -- node[anchor=west] {No} (step10);
\end{tikzpicture}

\caption{Flowchart illustrating the comparison of extracted attributes with the gold standard}

\label{fig:evaluation_flowchart}
\end{figure}
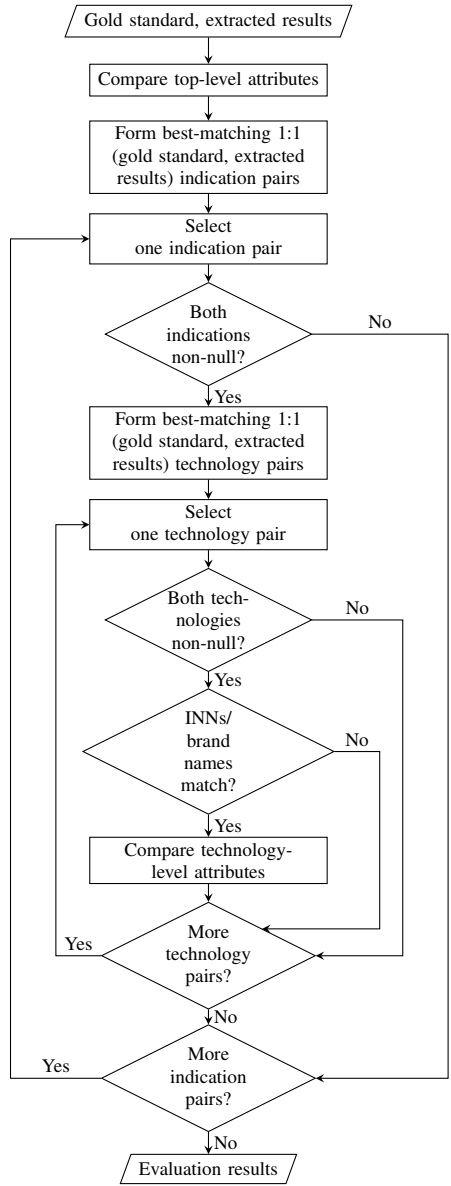

The evaluation process is illustrated in Figure \ref{fig:evaluation_flowchart}. We approach the problem of identifying the corresponding elements in the extracted and gold standard JSON objects by finding the path to each leaf-level element in one JSON object, and then finding the matching path in the other JSON object. To test if two paths are matching, we have to compare them at each nesting level. Since the values of the corresponding attributes at some level of the two paths might not be an exact lexical match, we also need to check for a semantic match. 

For non-leaf attributes which are lists of dictionaries, such as \texttt{indications} and \texttt{technologies}, matching a dictionary list element (a list element which is a dictionary) in the extracted results to the corresponding dictionary list element in the gold standard requires us to choose an identifier attribute (or anchor attribute) in the dictionary. An identifier attribute is a dictionary item that defines or identifies a dictionary list element in a list of dictionaries. It must be a defining characteristic of the parent attribute. At the \texttt{indications} level in the schema in Figure \ref{fig:schema}, the identifier attribute is \texttt{indication\_name}. At the \texttt{technologies} level, the primary identifier attribute is \texttt{inn} which corresponds to the INN (international nonproprietary name) of the drug. If no INN is present, \texttt{brand\_name} is used as the secondary identifier attribute. 

At each comparison step for a leaf-level attribute in Figure \ref{fig:evaluation_flowchart}, we first check if the attribute is present in both the extracted results and the gold standard. One of the situations listed in Table \ref{tab:classification} arises, and the match is classified as True Positive (TP), True Negative (TN), False Positive (FP), or False Negative (FN) according to this table.

\begin{table}[ht]
\centering
\caption{Classification of match between gold standard and extracted attributes}
\scriptsize
\begin{tabular}{|c|c|c|c|}
\hline
\textbf{Gold standard} & \textbf{Extracted} & \textbf{Values} & \textbf{Classification} \\
\textbf{attribute present?} & \textbf{attribute present?} & \textbf{match?} & \\
\hline
No & No & -- & TN \\
\hline
No & Yes & -- & FP \\
\hline
Yes & No & -- & FN \\
\hline
\multirow{2}{*}{Yes} & \multirow{2}{*}{Yes} & Yes & TP \\
\cline{3-4}
& & No & FP \\
\hline
\end{tabular}
\label{tab:classification}
\end{table}

When both the extracted and gold standard attributes are present, we first check if there is an exact match between the two values, using a simple string comparison. If this returns an exact match, we classify the result as TP. If the comparison does not result in an exact string match, we compare the two values using generative AI. We instruct the generative AI model to classify the result of the comparison into one of the following four categories:

\begin{enumerate}
\item \textit{Exact match:} The string comparison does not return an exact match, but the values are almost exactly the same, with very minor differences -- for example, an em dash is used instead of an en dash, or ``alpha" is used instead of $\alpha$. We classify this as TP.
\item \textit{Semantic match:} The values are lexically different but semantically identical, i.e., they convey the same meaning using different words, in the natural language sense of paraphrase \cite{bhagat2013} -- for example, if one of the values of \texttt{indication\_name} is ``pyrexia in children" and the other is ``fever in kids". We classify this as TP.
\item \textit{Useful match:} The values are neither lexically nor semantically identical, but the information extracted is still useful for the purposes of comparative HTA research -- for example, if one value of \texttt{indication\_name} is ``fever in children under 12 years of age" and the other is simply ``fever". We classify this as TP. Since HTA deals with reimbursement decisions and not clinical decisions, this classification is appropriate.
\item \textit{Non-match:} The values are neither lexically identical nor semantically similar. We classify this as FP.
\end{enumerate}

For the attributes which are lists of dictionaries (\texttt{indications} and \texttt{technologies}), in order to identify the corresponding dictionary list elements in the extracted data and the gold standard, we need to compare the corresponding identifier attributes. Multiple values of an identifier attribute in the two lists might match in some way (exact, semantic, or useful). We pick the best one-to-one match by ranking the matches by match type (exact $>$ semantic $>$ useful), and then choosing the match corresponding to the highest-ranked match type. In case there are multiple instances of the highest-ranked match type, we perform a greedy search \cite{black2005greedy} and choose the first match that we encounter.

\subsubsection{Model choice}
We chose the generative AI model Claude Opus 4.6 \cite{anthropic_opus46} for comparing the values when the comparison between the extracted results and the gold standard did not result in an exact string match. We chose this model because it was the most advanced model from Anthropic that was available at the time we performed the evaluation part of this study (March 2026). All the evaluations for the analyses listed in subsection \ref{subsec:analyses} were performed using Claude Opus 4.6, for the sake of uniformity of evaluation over different models, runs, and organisations.

\subsection{Analyses performed}
\label{subsec:analyses}

We performed the following analyses:

\begin{enumerate}
\item \textit{Performance metrics}: Computing performance metrics (precision, recall, accuracy, F1 score) for the extracted values, against the gold standard. 
\item \textit{Consistency analysis}: Testing the consistency of results in item 1 in this list over two runs with the same model, model parameters, prompt, and data. 
\item \textit{Time, costs, scalability}: Computing the time needed for extraction, the associated generative AI API costs, and how these quantities scale with number of documents.
\item \textit{Comparison with other methods}: Comparing with rule-based and traditional machine learning methods.
\item \textit{Generalisability}: Testing generalisability across different models.
\item \textit{Transferability}: Testing transferability across different HTA organisations and languages.
\end{enumerate}

Table \ref{tab:analyses} shows an overview of the analyses performed, and the results of these analyses are reported in Section \ref{sec:results}.

\begin{table*}[htbp]
\caption{Overview of analyses, with models used for extraction and evaluation, and the dates on which these were performed}
\label{tab:analyses}
\centering
\scriptsize
\renewcommand{\arraystretch}{1.2}
\begin{tabular}{|>{\raggedright\arraybackslash}p{3.2cm}|
                 >{\raggedright\arraybackslash}p{2.5cm}|
                 >{\raggedright\arraybackslash}p{5.5cm}|
                 >{\raggedright\arraybackslash}p{4cm}|}
\hline
\textbf{Analysis} & \textbf{HTA organisation} & \textbf{Extraction model (date)} & \textbf{Evaluation model (date)} \\
\hline
Performance metrics & NICE & Claude Opus 3 (July 2025) & Claude Opus 4.6 (March 2026) \\
Consistency analysis & NICE & Claude Opus 3 (July 2025) & Claude Opus 4.6 (March 2026) \\
Time, costs, scalability & NICE & Claude Opus 3 (July 2025) & N/A (only extraction time and costs compared) \\
Comparison with other methods & NICE & Rule-based, machine learning (2024-2025) & Evaluation performed manually (2024-2025)\\
Generalisability & NICE & Claude Opus 3, Claude Opus 4.6, Gemini 3.1 Pro, GPT OSS 120B (March 2026) & Claude Opus 4.6 (March 2026) \\
Transferability & NICE, ZIN, HAS & Claude Opus 4.6 (March 2026) & Claude Opus 4.6 (March 2026) \\
\hline
\end{tabular}
\end{table*}

\section{Results}
\label{sec:results}

\subsection{Performance metrics}
\label{performance}

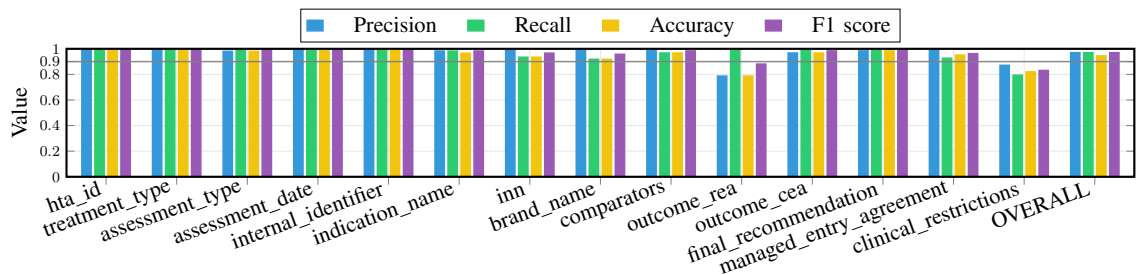
\begin{figure*}[htbp]
\centering
\begin{tikzpicture}[
    scale=0.7,
    transform shape
    ]
\begin{axis}[
    ybar,
    bar width=5pt,
    width=1.2\textwidth,
    height=4cm,
    symbolic x coords={
        hta\_id,
        treatment\_type,
        assessment\_type,
        assessment\_date,
        internal\_identifier,
        indication\_name,
        inn,
        brand\_name,
        comparators,
        outcome\_rea,
        outcome\_cea,
        final\_recommendation,
        managed\_entry\_agreement,
        clinical\_restrictions,
        OVERALL
    },
    xtick=data,
    xtick pos=bottom,
    x tick label style={
        rotate=20,
        anchor=east,
        font=\large
    },
    ylabel={Value},
    ylabel style={font=\large},
    ymin=0,
    ymax=1.0,
    ytick={0,0.2,0.4,0.6,0.8,0.9,1.0},
    yticklabel style={font=\footnotesize},
    legend style={
        at={(0.5,1.05)},
        anchor=south,
        font=\large,
        legend columns=4,
        /tikz/column sep=0.5cm
    },
    grid=major,
    grid style={gray!20},
    every axis plot/.append style={thick},
    enlarge x limits=0.04,
    clip=false,
    legend image code/.code={\draw[#1] (0cm,-0.1cm) rectangle (0.3cm,0.1cm);},
]
\addplot[
    fill=precisionblue,
    draw=precisionblue
] coordinates {
    (hta\_id, 1.0)
    (treatment\_type, 1.0)
    (assessment\_type, 0.98)
    (assessment\_date, 1.0)
    (internal\_identifier, 1.0)
    (indication\_name, 0.9824561403508771)
    (inn, 1.0)
    (brand\_name, 1.0)
    (comparators, 1.0)
    (outcome\_rea, 0.7868852459016393)
    (outcome\_cea, 0.9672131147540983)
    (final\_recommendation, 0.9836065573770492)
    (managed\_entry\_agreement, 1.0)
    (clinical\_restrictions, 0.8709677419354839)
    (OVERALL, 0.9699042407660738)
};
\addplot[
    fill=recallgreen,
    draw=recallgreen
] coordinates {
    (hta\_id, 1.0)
    (treatment\_type, 1.0)
    (assessment\_type, 1.0)
    (assessment\_date, 1.0)
    (internal\_identifier, 1.0)
    (indication\_name, 0.9824561403508771)
    (inn, 0.9344262295081968)
    (brand\_name, 0.9180327868852459)
    (comparators, 0.9672131147540983)
    (outcome\_rea, 1.0)
    (outcome\_cea, 1.0)
    (final\_recommendation, 1.0)
    (managed\_entry\_agreement, 0.926829268292683)
    (clinical\_restrictions, 0.7941176470588235)
    (OVERALL, 0.9699042407660738)
};
\addplot[
    fill=accuracyyellow,
    draw=accuracyyellow
] coordinates {
    (hta\_id, 1.0)
    (treatment\_type, 1.0)
    (assessment\_type, 0.98)
    (assessment\_date, 1.0)
    (internal\_identifier, 1.0)
    (indication\_name, 0.9655172413793104)
    (inn, 0.9344262295081968)
    (brand\_name, 0.9180327868852459)
    (comparators, 0.9672131147540983)
    (outcome\_rea, 0.7868852459016393)
    (outcome\_cea, 0.9672131147540983)
    (final\_recommendation, 0.9836065573770492)
    (managed\_entry\_agreement, 0.9508196721311475)
    (clinical\_restrictions, 0.819672131147541)
    (OVERALL, 0.9447236180904522)
};
\addplot[
    fill=f1purple,
    draw=f1purple
] coordinates {
    (hta\_id, 1.0)
    (treatment\_type, 1.0)
    (assessment\_type, 0.98989898989899)
    (assessment\_date, 1.0)
    (internal\_identifier, 1.0)
    (indication\_name, 0.9824561403508771)
    (inn, 0.9661016949152543)
    (brand\_name, 0.9572649572649572)
    (comparators, 0.9833333333333333)
    (outcome\_rea, 0.8807339449541284)
    (outcome\_cea, 0.9833333333333333)
    (final\_recommendation, 0.9917355371900827)
    (managed\_entry\_agreement, 0.9620253164556963)
    (clinical\_restrictions, 0.8307692307692308)
    (OVERALL, 0.9699042407660738)
};
\legend{Precision, Recall, Accuracy, F1 score}
\draw[gray, thick, opacity=0.9] ({rel axis cs:0,0.9}) -- ({rel axis cs:1,0.9});
\end{axis}
\draw[black, thick] (current axis.south west) rectangle (current axis.north east);
\end{tikzpicture}
\vspace{-15pt}
\caption{Evaluation metrics per attribute, and over all attributes}
\label{fig:hta_evaluation_metrics}
\end{figure*}

Figure \ref{fig:hta_evaluation_metrics} shows the performance metrics (precision, recall, accuracy, F1 score) per attribute, and also over all attributes. We see that for 12 of the 14 attributes extracted from HTA documents, all the metrics have a value $> 90\%$. For two attributes, \texttt{outcome\_rea} and \texttt{clinical\_restrictions}, the performance is poorer -- the F1 score is 0.88 for \texttt{outcome\_rea} and 0.83 for \texttt{clinical\_restrictions}. 

Extracting the attribute \texttt{outcome\_rea} is a difficult task, even for a human. The relative effectiveness discussion in HTA documents is often long, nuanced, and ambiguous. When the document contains sentences along the lines of ``This drug seems to be clinically effective, but the evidence presented is insufficient", generative AI usually classified \texttt{outcome\_rea} as positive, while the human expert usually classified it as negative in the gold standard.

For the attribute \texttt{clinical\_restrictions}, the reason for the relatively poor performance is the fact that this attribute is reported along with the attribute \texttt{indication\_name}, in the same section of the HTA document -- the ``Recommendations" section, making it difficult to decouple \texttt{clinical\_restrictions} from \texttt{indication\_name}. As a toy example, suppose that a drug is being assessed for treating fever in children under 12 years of age. The human expert might extract the two attributes as \{\texttt{indication\_name}: ``fever in children under 12 years of age", \texttt{clinical\_restrictions}: \texttt{null}\}, which will be the gold standard. Generative AI might extract the attributes as \{\texttt{indication\_name}: ``fever", \texttt{clinical\_restrictions}: ``only for children under 12 years of age"\}. Since we also accept useful matches during the evaluation process, the extracted \texttt{indication\_name} will be considered as matching the gold standard and the match will be classified as TP, but the extracted \texttt{clinical\_restrictions} will not match the gold standard and the match will be classified as FP.

During the evaluation process, most (634 out of 796) of our comparisons against the gold standard yielded an exact match during the initial check for an exact string match, which is a rule-based, deterministic check. For the comparisons that did not yield an exact string match, we used the generative AI model Claude Opus 4.6 to classify the result of the comparison between the extracted and gold standard values as an exact match, semantic match, useful match, or non-match. 

We followed up the comparisons that were performed by generative AI with a manual check to ascertain if the AI had classified the result of each comparison correctly. Claude Opus 4.6 found 20 exact matches, 19 semantic matches, 84 useful matches, and 39 non-matches. The human and the AI agreed on all the non-matches. For the other matches that were classified as exact, semantic, or useful matches by the AI, the human sometimes classified them differently, but still as one of these three kinds of matches -- not as a non-match. Since a non-match is classified as FP and exact, semantic, and useful matches are all classified as TP, this discrepancy is immaterial for the purposes of calculating the metrics. Thus we verified that the human and the AI agreed on the performance metrics, even though there was some discrepancy between the two regarding exact/semantic/useful match classification.

\subsection{Consistency analysis}

Due to the generative variability of generative AI models, the results from two runs with the same model, model parameters, prompt, and data may not necessarily be the same \cite{lee2024one}. To examine the consistency (in the sense of test-retest reliability \cite{Vilagut2014}) of our results over different runs, we performed two runs, and used the results from the first run as the gold standard for the second run. We calculated the performance metrics, and the values of all the metrics for all the attributes were $>96\%$. 

\subsection{Time, costs, scalability}
On average, it took Claude Opus 3 $\sim$0.5 minute to extract the attributes from one HTA document, and the API cost was $\sim$0.5 euro per document. For comparison, it takes a domain expert $\sim$15 minutes to extract the attributes from one document. Thus there is a $\sim$30-fold increase in time-efficiency when generative AI is employed instead of a human for this task. The time and API costs scale linearly with the number of documents. 

\subsection{Comparison with other methods}
\label{baselines}

We compared our schema-based generative AI method to two traditional methods -- rule-based and machine learning approaches. The details of the comparison are presented in a companion paper focused on the analysis and implications of these three approaches for comparative HTA research \cite{versteeg2026hta}. Here we present a summary of the results.

Since no existing traditional method that we are aware of can extract data which has the complex structure illustrated in Figure \ref{fig:schema}, in a single call, we do not have other methods to directly compare with, in terms of the entire output containing all the attributes. However, we could compare our method with existing traditional methods a few attributes at a time. 

Using a rule-based method, we were able to extract four attributes -- \texttt{hta\_id}, \texttt{assessment\_date}, \texttt{internal\_identifier}, and \texttt{indication\_name} with \>90\% accuracy, and \texttt{assessment\_type}, \texttt{inn}, and \texttt{brand\_name} with  \>70\% accuracy. We were not able to extract the other attributes with significant accuracy.

We tried several machine learning algorithms to classify the three attributes whose values could be broadly described as binary (positive/negative) -- \texttt{outcome\_rea}, \texttt{outcome\_cea}, and \texttt{final\_recommendation}. The methods we tried were Gradient Boosting, Support Vector Machine, Logistic Regression, Random Forest, Naive Bayes, and BERT. For the \texttt{outcome\_rea} attribute, the best-performing models were Gradient Boosting and Random Forest with an F1 score of 0.76 for both models. For the \texttt{outcome\_cea} and \texttt{final\_ recommendation} attributes, the best-performing model was Gradient Boosting with F1 scores of 0.98 and 0.93 respectively. 

Comparing these metrics to those in Figure \ref{fig:hta_evaluation_metrics}, we see that the generative AI approach outperforms both rule-based and machine learning approaches. In addition, this method can extract all the desired attributes from a document via a single call to the model. This makes the implementation simpler as compared to both the rule-based method, which needs the creation of a separate rule for each attribute, and the machine-learning methods, which need labelled training data for each attribute to be classified. 

\subsection{Generalisability across models}
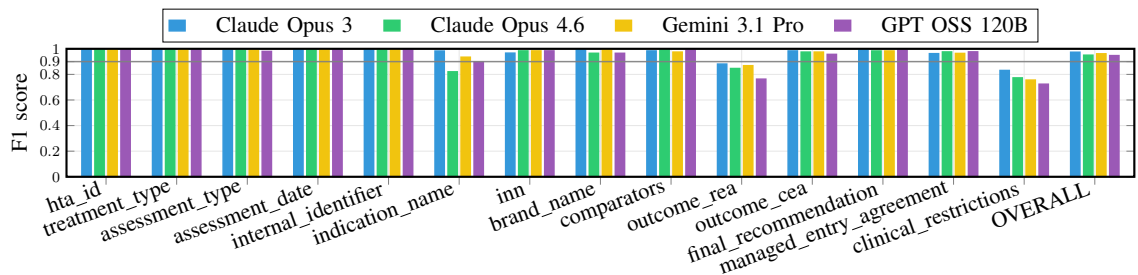
\begin{figure*}[htbp]
\centering
\begin{tikzpicture}[
    scale=0.7,
    transform shape
    ]
\begin{axis}[
    ybar,
    bar width=5pt,
    width=1.2\textwidth,
    height=4cm,
    symbolic x coords={
        hta\_id,
        treatment\_type,
        assessment\_type,
        assessment\_date,
        internal\_identifier,
        indication\_name,
        inn,
        brand\_name,
        comparators,
        outcome\_rea,
        outcome\_cea,
        final\_recommendation,
        managed\_entry\_agreement,
        clinical\_restrictions,
        OVERALL
    },
    xtick=data,
    xtick pos=bottom,
    x tick label style={
        rotate=20,
        anchor=east,
        font=\large
    },
    ylabel={F1 score},
    ylabel style={font=\large},
    ymin=0,
    ymax=1.0,
    ytick={0,0.2,0.4,0.6,0.8,0.9,1.0},
    yticklabel style={font=\footnotesize},
    legend style={
        at={(0.5,1.05)},
        anchor=south,
        font=\large,
        legend columns=4,
        /tikz/column sep=0.5cm
    },
    grid=major,
    grid style={gray!20},
    every axis plot/.append style={thick},
    enlarge x limits=0.04,
    clip=false,
    legend image code/.code={\draw[#1] (0cm,-0.1cm) rectangle (0.3cm,0.1cm);},
]
\addplot[
    fill=precisionblue,
    draw=precisionblue
] coordinates {
    (hta\_id, 1.0)
    (treatment\_type, 1.0)
    (assessment\_type, 0.98989898989899)
    (assessment\_date, 1.0)
    (internal\_identifier, 1.0)
    (indication\_name, 0.9824561403508771)
    (inn, 0.9661016949152543)
    (brand\_name, 1.0)
    (comparators, 0.9833333333333333)
    (outcome\_rea, 0.8807339449541284)
    (outcome\_cea, 0.9833333333333333)
    (final\_recommendation, 0.9917355371900827)
    (managed\_entry\_agreement, 0.9620253164556963)
    (clinical\_restrictions, 0.8307692307692308)
    (OVERALL, 0.973232669869595)
};
\addplot[
    fill=recallgreen,
    draw=recallgreen
] coordinates {
    (hta\_id, 1.0)
    (treatment\_type, 1.0)
    (assessment\_type, 0.98989898989899)
    (assessment\_date, 1.0)
    (internal\_identifier, 1.0)
    (indication\_name, 0.8201438848920863)
    (inn, 0.9836065573770492)
    (brand\_name, 0.9649122807017544)
    (comparators, 1.0)
    (outcome\_rea, 0.846153846153846)
    (outcome\_cea, 0.9743589743589743)
    (final\_recommendation, 0.983050847457627)
    (managed\_entry\_agreement, 0.975609756097561)
    (clinical\_restrictions, 0.7727272727272727)
    (OVERALL, 0.9500998003992017)
};
\addplot[
    fill=accuracyyellow,
    draw=accuracyyellow
] coordinates {
    (hta\_id, 1.0)
    (treatment\_type, 1.0)
    (assessment\_type, 0.98989898989899)
    (assessment\_date, 1.0)
    (internal\_identifier, 1.0)
    (indication\_name, 0.9344262295081968)
    (inn, 0.9836065573770492)
    (brand\_name, 1.0)
    (comparators, 0.9743589743589743)
    (outcome\_rea, 0.8679245283018869)
    (outcome\_cea, 0.9743589743589743)
    (final\_recommendation, 0.983050847457627)
    (managed\_entry\_agreement, 0.963855421686747)
    (clinical\_restrictions, 0.7555555555555554)
    (OVERALL, 0.9609164420485176)
};
\addplot[
    fill=f1purple,
    draw=f1purple
] coordinates {
    (hta\_id, 1.0)
    (treatment\_type, 1.0)
    (assessment\_type, 0.9795918367346939)
    (assessment\_date, 1.0)
    (internal\_identifier, 1.0)
    (indication\_name, 0.896)
    (inn, 0.9836065573770492)
    (brand\_name, 0.9649122807017544)
    (comparators, 0.9915966386554621)
    (outcome\_rea, 0.7628865979381444)
    (outcome\_cea, 0.9565217391304348)
    (final\_recommendation, 0.9915966386554621)
    (managed\_entry\_agreement, 0.9761904761904763)
    (clinical\_restrictions, 0.7234042553191489)
    (OVERALL, 0.9468728984532616)
};
\legend{Claude Opus 3, Claude Opus 4.6, Gemini 3.1 Pro, GPT OSS 120B}
\draw[gray, thick, opacity=0.9] ({rel axis cs:0,0.9}) -- ({rel axis cs:1,0.9});
\end{axis}
\draw[black, thick] (current axis.south west) rectangle (current axis.north east);
\end{tikzpicture}
\vspace{-15pt}
\caption{F1 scores per attribute, and over all attributes, for the generative AI models Claude Opus 3, Claude Opus 4.6, Gemini 3.1 Pro, and GPT OSS 120B}
\label{fig:f1_score_comparison_genai_models}
\end{figure*}

We used this schema-based framework to extract results using three other models -- two proprietary models, Claude Opus 4.6 and Gemini 3.1 Pro \cite{google_gemini31pro}, and the open-weights model GPT OSS 120B \cite{openai2025gptoss} accessed through the AI-Hub \cite{surf_aihub} provided by SURF \cite{surf}, the Dutch IT cooperative for research and education. Claude Opus 4.6 was used as the model for the automated evaluation in all these cases. Figure \ref{fig:f1_score_comparison_genai_models} shows the F1 scores for these four models. It can be seen that we get comparable results from all four models, demonstrating the generalisability of this framework across different generative AI models.

\subsection{Transferability across HTA organisations and languages}

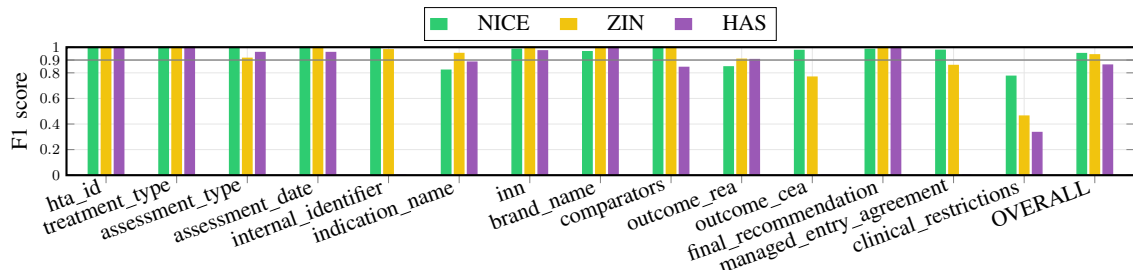
\begin{figure*}[htbp]
\centering
\begin{tikzpicture}[
    scale=0.7,
    transform shape
    ]
\begin{axis}[
    ybar,
    bar width=5pt,
    width=1.2\textwidth,
    height=4cm,
    symbolic x coords={
        hta\_id,
        treatment\_type,
        assessment\_type,
        assessment\_date,
        internal\_identifier,
        indication\_name,
        inn,
        brand\_name,
        comparators,
        outcome\_rea,
        outcome\_cea,
        final\_recommendation,
        managed\_entry\_agreement,
        clinical\_restrictions,
        OVERALL
    },
    xtick=data,
    xtick pos=bottom,
    x tick label style={
        rotate=20,
        anchor=east,
        font=\large
    },
    ylabel={F1 score},
    ylabel style={font=\large},
    ymin=0.0,
    ymax=1.0,
    ytick={0.0,0.2,0.4,0.6,0.8,0.9,1.0},
    yticklabel style={font=\footnotesize},
    legend style={
        at={(0.5,1.05)},
        anchor=south,
        font=\large,
        legend columns=3,
        /tikz/column sep=0.5cm
    },
    grid=major,
    grid style={gray!20},
    every axis plot/.append style={thick},
    enlarge x limits=0.04,
    clip=false,
    legend image code/.code={\draw[#1] (0cm,-0.1cm) rectangle (0.3cm,0.1cm);},
]
\addplot[
    fill=recallgreen,
    draw=recallgreen
] coordinates {
    (hta\_id, 1.0)
    (treatment\_type, 1.0)
    (assessment\_type, 0.98989898989899)
    (assessment\_date, 1.0)
    (internal\_identifier, 1.0)
    (indication\_name, 0.8201438848920863)
    (inn, 0.9836065573770492)
    (brand\_name, 0.9649122807017544)
    (comparators, 1.0)
    (outcome\_rea, 0.846153846153846)
    (outcome\_cea, 0.9743589743589743)
    (final\_recommendation, 0.983050847457627)
    (managed\_entry\_agreement, 0.975609756097561)
    (clinical\_restrictions, 0.7727272727272727)
    (OVERALL, 0.9500998003992017)
};
\addplot[
    fill=accuracyyellow,
    draw=accuracyyellow
] coordinates {
    (hta\_id, 1.0)
    (treatment\_type, 1.0)
    (assessment\_type, 0.9130434782608696)
    (assessment\_date, 1.0)
    (internal\_identifier, 0.9795918367346939)
    (indication\_name, 0.9508196721311476)
    (inn, 1.0)
    (brand\_name, 1.0)
    (comparators, 1.0)
    (outcome\_rea, 0.9056603773584906)
    (outcome\_cea, 0.7659574468085107)
    (final\_recommendation, 1.0)
    (managed\_entry\_agreement, 0.8571428571428571)
    (clinical\_restrictions, 0.4615384615384615)
    (OVERALL, 0.9401459854014599)
};
\addplot[
    fill=f1purple,
    draw=f1purple
] coordinates {
    (hta\_id, 1.0)
    (treatment\_type, 1.0)
    (assessment\_type, 0.9583333333333334)
    (assessment\_date, 0.9583333333333334)
    (internal\_identifier, 0)
    (indication\_name, 0.8831168831168832)
    (inn, 0.9714285714285714)
    (brand\_name, 1.0)
    (comparators, 0.8421052631578947)
    (outcome\_rea, 0.9032258064516129)
    (outcome\_cea, 0)
    (final\_recommendation, 1.0)
    (managed\_entry\_agreement, 0)
    (clinical\_restrictions, 0.33333333333333337)
    (OVERALL, 0.8601190476190477)
};
\legend{NICE, ZIN, HAS}
\draw[gray, thick, opacity=0.9] ({rel axis cs:0,0.9}) -- ({rel axis cs:1,0.9});
\end{axis}
\draw[black, thick] (current axis.south west) rectangle (current axis.north east);
\end{tikzpicture}
\vspace{-15pt}
\caption{F1 scores per attribute, and over all attributes, for the HTA organisations NICE, ZIN, and HAS. HAS does not report the attributes \texttt{internal\_identifier} and \texttt{outcome\_cea}. For \texttt{managed\_entry\_agreement} for HAS, either both the AI-extracted value and the gold standard value were \texttt{null}, or the AI did not classify the requirement of future assessments as an outcome-based managed entry agreement like the human expert did.}
\label{fig:f1_score_comparison_hta_orgs}
\end{figure*}

In Figure \ref{fig:f1_score_comparison_hta_orgs} we compare the performance of our pipeline designed for NICE documents, as-is, without modification, on 25 documents each from Zorginstituut Nederland (ZIN) \cite{ZIN}, the Dutch HTA organisation, and Haute Autorité de Santé (HAS) \cite{HAS}, the French HTA organisation. Claude Opus 4.6 was used for both the extraction and evaluation steps. We see that our approach performs reasonably well for several attributes for ZIN and HAS too, thus illustrating the broad transferability of our schema-based framework across different HTA organisations and languages.

\section{Conclusions and discussion}
\label{sec:conclusions}

Using a schema-based framework, we have successfully extracted data with a hierarchical structure, containing attributes of variable cardinality, from HTA documents. We have also evaluated the results in an automated manner. We used a single, unified schema and generative AI in both the extraction and evaluation
steps.

While we have integrated the extraction and evaluation steps together in the single end-to-end pipeline shown in Figure \ref{fig:extraction-pipeline}, the evaluation step is independent of the extraction step. The evaluation algorithm can be applied to any outputs that follow a prescribed schema, regardless of the extraction method, provided the schema and a gold standard which follows the same schema are available. Since the evaluation step needs a gold standard, the evaluation algorithm can only be used on the labelled development and test datasets, and not in production, on unseen data.

Since we developed our prompt on a development set for NICE documents, using the generative AI model Claude Opus 3, we get the best performance for this combination of HTA organisation and generative AI model. While still broadly performing well, differences in the way different HTA organisations report assessments, as well as model differences between model providers and model drift between generations of models from the same provider, might cause the performance to degrade. Refining the prompt based on documents published by a particular HTA organisation, using a particular generative AI model, might improve performance. 

The apparently poorer performance of Claude Opus 4.6, a more advanced model than Claude Opus 3, is in some cases due to the fact that Claude Opus 4.6 extracts the attributes at a higher degree of granularity than either the gold standard or the data extracted by Claude Opus 3. Thus it might be necessary to make the prompt more specific and explicitly specify the level of granularity desired in the output.

We use a proprietary generative AI model, Claude Opus 3, for this information extraction task. We do not have access to the model itself, and only use it via its public API. Thus the availability, latency, and costs depend on the proprietary model provider and these parameters can change in the future. Of particular concern is the fact that proprietary model providers routinely deprecate older models in favour of new ones, which endangers not only the reproducibility of the results but also the long-term functionality and sustainability of the information extraction and evaluation pipeline. 

With open-weights models increasing in size and capability, the problems associated with the use of proprietary models can in principle be avoided by downloading and locally deploying state-of-the-art open-weights models. Figure \ref{fig:f1_score_comparison_genai_models} shows that the performance of the open-weights model GPT OSS 120B is comparable to that of state-of-the-art proprietary models. Thus this method can be used for information extraction from private and/or sensitive documents, provided an organisation possesses the technical expertise and infrastructure for hosting such open-weights models locally.

The gold standard dataset that we used was created by a single domain expert. Manual extraction of attributes by humans is itself prone to subjectivity and error \cite{Wang2020}, thus, for a more robust assessment of the performance of this method, one should obtain gold standard datasets created by multiple domain experts for the same set of HTA documents, and perform inter-rater reliability analysis \cite{interrater_reliability, mitra2024ieee} to compare the agreement between multiple human raters, and that between human raters and AI. 

While we have used generative AI to automate the tedious extraction and evaluation steps, human expertise was indispensable in this study. HTA domain experts provided the domain knowledge about the attributes to extract and their structure, and they prepared the gold standards. The Research Engineers designed the schema and wrote the scaffolding code in Python to programmatically extract the attributes via the generative AI API. The match classifications performed by generative AI during the evaluation step were manually verified by humans. Thus human expertise and involvement remain essential when the results need to be highly valid \cite{software_accuracy_reliability, creswell2023research}, like in this use case -- comparative HTA research.

\section{Future work}
\label{sec:future_work}

We are using this schema-based framework for extracting complex information in two domains very different from HTA -- a scoping review of young people's geographies, which is from the human geography domain, and extracting sociodemographic attributes from the Oxford Dictionary of National Biography \cite{odnb}, which is from the historical sociology domain. The preliminary results from these two projects are promising.

\section*{Acknowledgement}
We thank Martine de Vos and Ingo Schr\"oder for reading the manuscript and providing feedback that improved the paper. 

\textit{Generative AI use disclosure:} Generative AI was used for coding assistance to write the pipeline, and to generate the \LaTeX{} code for the tables, figures, and references in this paper. The generative AI tools used were OpenAI's GPT series of models, Anthropic's Claude series of models, and Google's Gemini series of models. All the content was reviewed carefully by the authors and the authors take full responsibility for the final content of the manuscript.

\bibliographystyle{IEEEtran}
\bibliography{HTA}

\end{document}